\documentclass[letterpaper, 10 pt, conference]{ieeeconf}  
\usepackage[style=ieee,maxnames=3,minnames=1]{biblatex}
\usepackage{array}
\usepackage{multirow}
\usepackage{booktabs}
\usepackage{pifont}
\usepackage{multicol}
\usepackage{wrapfig}
\usepackage{float}
\usepackage{afterpage}
\usepackage{stfloats}
\usepackage{tabularx}
\usepackage{array}
\usepackage{makecell}
\usepackage{multirow}
\usepackage{dsfont}

\usepackage{mwe} 
\usepackage[utf8]{inputenc} 
\usepackage[T1]{fontenc}    
\usepackage{hyperref}       
\usepackage{url}     
\usepackage{subcaption}
\usepackage{graphicx}
\usepackage{booktabs}       
\usepackage{amsfonts}       
\usepackage{nicefrac}       
\usepackage{array}
\usepackage{amsmath}
\usepackage{microtype}      
\usepackage{xcolor}         
\usepackage{subcaption}
\usepackage{tablefootnote}
\newcolumntype{C}[1]{>{\centering\arraybackslash}p{#1}}

\renewcommand{\eqref}[1]{Eqn~\ref{#1}}
\newcommand{\figref}[1]{Figure~\ref{#1}}

\newcommand{\tabref}[1]{Table~\ref{#1}}

\newcommand{\ie}{\textrm{i.e.}}
\newcommand{\eg}{\textrm{e.g.}}

\newcommand{\bbone}{\text{\usefont{U}{bbold}{m}{n}1}}
\MakeRobust{\bbone}

\newcommand{\openvla}{OpenVLA}
\newcommand{\pizero}{$\pi_0$}

\newcommand{\ouropenvla}{\openvla{}$^+$}
\newcommand{\ourpizero}{\pizero{}$^+$}

\newcommand{\pickcan}{\texttt{PickCan}}
\newcommand{\opendrawer}{\texttt{OpenDrawer}}
\newcommand{\closedrawer}{\texttt{CloseDrawer}}
\newcommand{\movenear}{\texttt{MoveNear}}
\newcommand{\pickcarrot}{\texttt{PickCarrot}}
\newcommand{\pickeggplant}{\texttt{PickEggplant}}
\newcommand{\pickknife}{\texttt{PickKnife}}
\newcommand{\placeknife}{\texttt{PlaceKnifeNearCorn}}
\newcommand{\putknifecloth}{\texttt{PutKnifeOnCloth}}
\newcommand{\putcarrotplate}{\texttt{PutCarrotOnPlate}}
\newcommand{\placecornknife}{\texttt{PlaceCornNearKnife}}

\newcommand{\putcarrotcolorplate}{\texttt{PutCarrotOn\{Color\}Plate}}

\IEEEoverridecommandlockouts                              

\title{\LARGE \bf
Enhancing Generalization in Vision–Language–Action Models by Preserving Pretrained Representations
}

\author{
\authorblockN{Shresth Grover$^{1}$\authorrefmark{1},
Akshay Gopalkrishnan$^{1}$\authorrefmark{1},
Bo Ai$^{1}$,
Henrik I. Christensen$^{1}$,
Hao Su$^{1,2}$, and Xuanlin Li$^{2}$}
\vspace{2pt}
\authorblockA{$^{1}$UC San Diego \quad $^{2}$Hillbot \quad \authorrefmark{1}Equal contribution}
\vspace{2pt}
\authorblockA{\textbf{\textcolor{magenta}{\url{https://gen-vla.github.io}}}}
\vspace{-15pt}
}

\begin{document}

\maketitle
\begin{abstract}
Vision–language–action (VLA) models finetuned from vision–language models (VLMs) hold the promise of leveraging rich pretrained representations to build generalist robots across diverse tasks and environments. However, direct fine-tuning on robot data often disrupts these representations and limits generalization. We present a framework that better preserves pretrained features while adapting them for robot manipulation. Our approach introduces three components: (i) a dual-encoder design with one frozen vision encoder to retain pretrained features and another trainable for task adaptation, (ii) a string-based action tokenizer that casts continuous actions into character sequences aligned with the model’s pretraining domain, and (iii) a co-training strategy that combines robot demonstrations with vision–language datasets emphasizing spatial reasoning and affordances. Evaluations in simulation and on real robot show that our method improves robustness to visual perturbations, generalization to novel instructions and environments, and overall task success compared to baselines. 
\end{abstract}

\section{Introduction}
Building general-purpose robots that generalize across tasks and environments is a long-standing goal in robotics. While foundation models in vision and language generalize well from internet-scale data~\cite{Bommasani2021FoundationModels}, the lack of large-scale, action-labeled robot data makes comparable generalization in robotic manipulation challenging~\cite{Ken2025good}. Efforts on vision-language-action (VLA) models seek to improve robotic generalization by leveraging the rich representations of pretrained vision-language models (VLMs) and finetuning them on robot data (\eg, \cite{Black2024pizero, Kim2024openvla}). This allows us to leverage advances in VLMs for robotic tasks, presenting a promising pathway toward manipulation policies that generalize across environments, tasks, and embodiments. 

\begin{figure*}[t]
    \centering
    \includegraphics[width=1.0\linewidth]{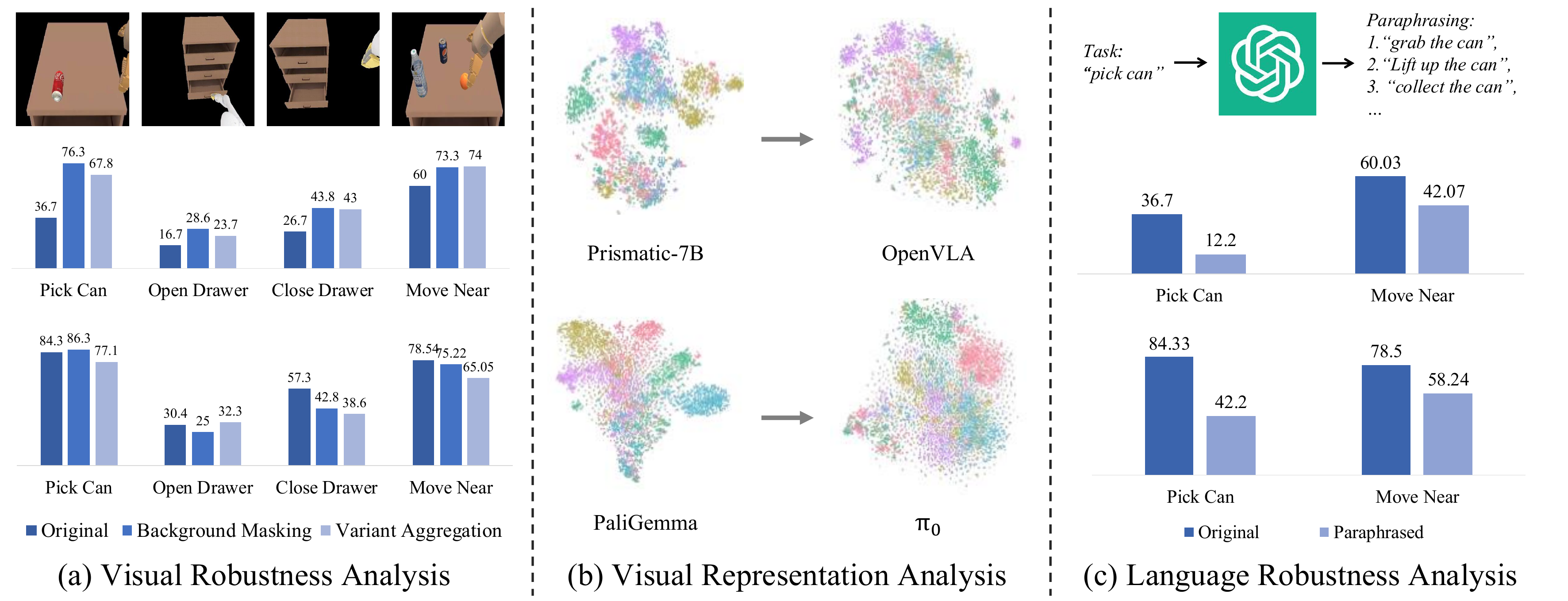}
    \caption{
    \textbf{Motivating experiments.} (a) Performance for both OpenVLA (top) and $\pi_0$ (down) is significantly impacted by background variations (\eg, background masking and randomization), indicating visual overfitting. (b) t-SNE plots of visual encoders show that pretrained visual representations deteriorate when VLA models are trained by directly finetuning VLMs on action data. (c) Success rates for both OpenVLA and $\pi_0$ decrease even with slightly-rephrased instructions, revealing language overfitting. While text augmentation helps, the performance still lags significantly behind our proposed approach.}
    \label{fig:motivating}
\end{figure*}

However, directly finetuning pretrained VLMs on robot data leads to significant representation degradation. Our preliminary experiments show that, with an existing training recipe \cite{Kim2024openvla,Black2024pizero}, background changes and small instruction paraphrases can cause large drops in performance (\figref{fig:motivating}), suggesting that finetuning disrupts the structures of pretrained visual and language representations. Recent work has proposed co-training on both vision-language and robotic data~\cite{yang2025magma,intelligence2025pi05} to mitigate this issue. However, we find that naive co-training on both objectives does not lead to the best performance, as the two sources of data differ substantially in structure, limiting the ability for the robot action prediction process to reuse vision-language representations. These findings point to an important question: what is the training recipe that preserves the powerful, general representations of pretrained VLMs to facilitate generalization of downstream VLAs? 

To address this challenge, we explore three design choices. First, we propose mixing frozen and finetuned visual encoders to preserve pretrained VLM representations while keeping high model flexibility to adapt well to robotic tasks. Second, we propose a language-aligned action tokenizer that casts numerical robot actions into character sequences, thereby enabling maximal reuse of pretrained language representations and allowing actions to be refined step by step during generation. Finally, leveraging this unified string-based output space, we co-train the model on both robot and vision-language datasets that emphasize spatial affordances and reasoning. The training recipe is general and can be applied to different existing VLA model architectures.


We evaluate these designs in both simulation and real-world settings and find consistent improvements over baseline VLAs. Our approach yields stronger generalization to out-of-distribution visuals, such as backgrounds, table textures, and distractor objects, as well as to varied language commands, leading to higher task success rates.
These results provide practical insights into how pretrained VLMs can be effectively grounded in robotic action, bringing us closer to reliable generalist robot manipulation policies.

\section{Related Work}

\textbf{Vision-Language Models (VLMs).}  
VLMs learn unified representations across vision and language modalities through large-scale pretraining on web-scale image-text data. Recent VLMs such as PaliGemma~\cite{beyer2024paligemma}, Qwen2.5-VL~\cite{Bai2025Qwen2.5VL}, InternVL3~\cite{zhu2025internvl3exploring}, and many others~\cite{wang2024qwen2,li2024llava,li2023blip,zhu2023minigpt,karamcheti2024prismatic,Deitke2024Molmo,Lin2024VILA,Abdin2024Phi3,Li2024LLaVAOneVision,Hong2024CogVLM2,Yao2024MiniCPMV} combine strong language models with pretrained visual encoders to achieve generalization across diverse tasks such as image captioning, visual-question answering, and spatial reasoning. Visual backbones like DINOv2 \cite{Oquab2024Dinov2} and SigLIP \cite{Zhai2023Sigmoid} provide robust spatial and semantic grounding, while language models such as Gemma~\cite{Mesnard2024Gemma}, LLaMA \cite{touvron2023llama}, and many others \cite{Chowdhery2023PALM, DeepSeek2024DeepSeekV3, DeepSeek2025DeepSeekR1, Gunasekar2023Textbooks, Abdin2024Phi3, OpenAI2023GPT4, Jiang2024Mixtral} enable instruction following and compositional reasoning. While these models show impressive zero-shot and few-shot capabilities, they are primarily optimized for passive perception and reasoning tasks. Adapting them for embodied agents introduces challenges in preserving spatial grounding capabilities, visual robustness, and cross-modal alignment when transitioning from static vision-language data to dynamic robotic control.

\textbf{Vision-Language-Action Models (VLAs).}  
VLAs~\cite{li2025hamster, wen2025dexvla, cui2025openhelix, cheang2024gr2, Ghosh2024Octo, Shridhar2022PerAct, OpenX, Goyal2024RVT2, Jiang2022VIMA, liu2025hybridvla, bjorck2025gr00t} adapt pretrained VLMs for robotic control by fine-tuning them on action-labeled datasets. A widely adopted approach in works like OpenVLA and Magma~\cite{zitkovich2023rt, Kim2024openvla} remaps rarely used language tokens to bins of discretized actions, discarding their pretrained semantics and breaking alignment with the original vision-language representations. Fine-tuning exclusively on robotic data also leads to degradation of visual representations, reducing robustness to distractors, viewpoint shifts, or rephrased instructions. Several methods attempt to mitigate this via input augmentation~\cite{xiaoaugmentation, Yu2023Scaling}, spatially enriched encoders~\cite{Radford2021Learning, Oquab2024Dinov2, Zhai2023Sigmoid, Shang2024Theia}, or auxiliary cues like end-effector traces~\cite{Gu2024RTTrajectory, Niu2024LLARVA}, but their training data is still limited to robotic domains. Recent works have also explored co-training on both vision-language and action data~\cite{yang2025magma,li2025hamster,intelligence2025pi05}, but it is found that robotic action prediction and vision-language reasoning often involve conflicting training objectives, and naïvely co-training on both data types can degrade performance in each domain~\cite{yang2025magma}. In this work, we address these co-training issues with a partially frozen visual encoder designed to better preserve pretrained robust visual representations during fine-tuning, and an action tokenizer that reuses pretrained embeddings aligned better with the vision-language training objective. 

\begin{figure*}[t]
    \centering
    \includegraphics[width=\linewidth]{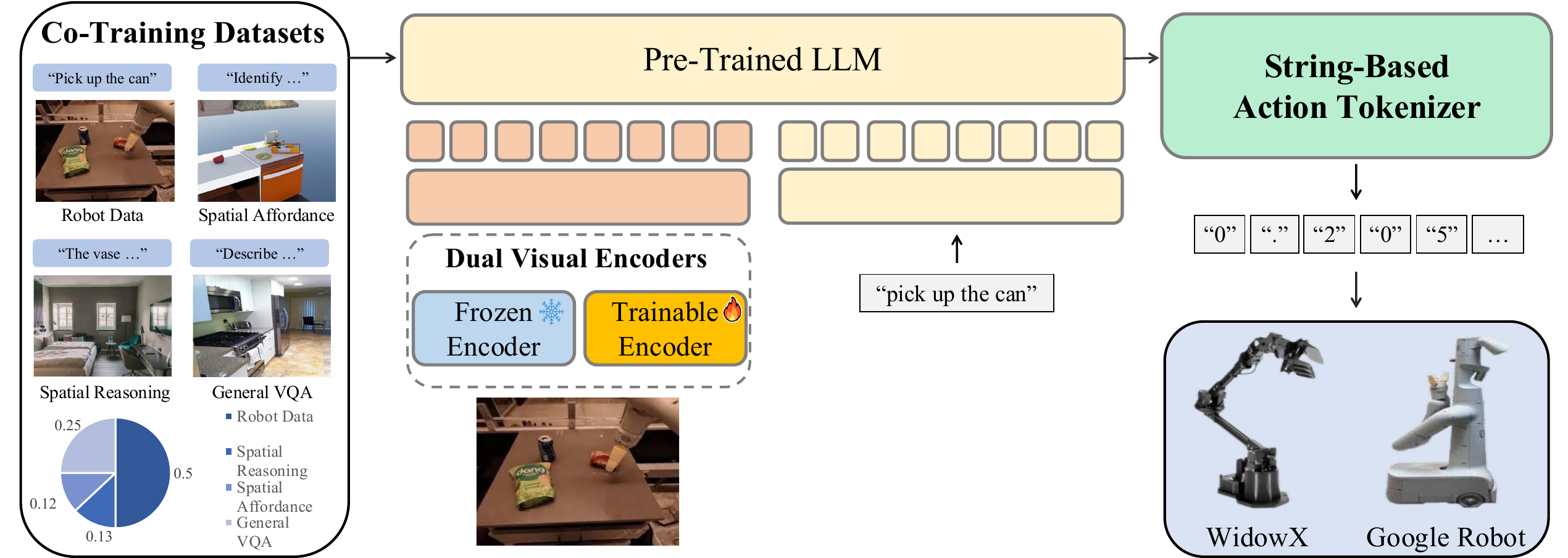}
    \caption{
    \textbf{Method overview.} Our approach improves VLA generalization through three designs:
    (a)~\textbf{Co-training:} Jointly training on robotic and vision-language datasets that emphasize spatial affordance and reasoning helps preserve pretrained representations for generalization.
    (b)~\textbf{Partially-frozen visual encoders:} One encoder is frozen to retain robust pretrained features from VLMs, whereas the other keeps the full flexibility to specialize for the robot tasks.
    (c)~\textbf{String tokenizer:} Robotic actions are expressed as digit-based strings to maximally reuse pretrained language representations and to unify prediction targets across non-robotic and robotic task domains for co-training.
    }
    \label{fig:main_fig2}
\end{figure*}

\section{Method}
\label{sec:method}

The robotic manipulation problem can be modeled as a partially observable Markov decision process (POMDP)~\cite{ai2022deep, leslie1998planning}. At the beginning of each episode, the agent receives a natural language command $c \in \mathcal{C}$, which remains fixed for the duration of the task. At each timestep $t$, the agent observes $o_t \in \mathcal{O}$, typically an RGB image, and predicts a short-horizon trajectory $\tau_t = (a_t, \ldots, a_{t+H-1})$ consisting of $H$ low-level actions. Each action $a_t \in \mathcal{A}$ is parameterized as
\[
a_t = (\Delta x, \Delta y, \Delta z, \phi, \theta, \psi, g),
\]
where $(\Delta x, \Delta y, \Delta z)$ and $(\phi, \theta, \psi)$ denote the end-effector’s translation and rotation, and $g \in \{0, 1\}$ indicates the gripper state (open or closed). The policy $\pi$ thus maps observations and commands to action trajectories,
optimized to minimize a task-specific objective,
\[
\hat{\tau}_t = \arg\min_{\tau_t} \; \mathcal{L}(\tau_t, \tau_t^*),
\]
where $\tau_t^*$ is the expert trajectory. 

Directly fine-tuning pretrained VLMs for this objective often results in representation collapse and limited generalization, as shown in Fig.~\ref{fig:motivating}. To address this, we introduce three components that can be integrated into the training process of many different VLA models.

\subsection{Partially-Frozen Dual Encoder Architecture}

We first study whether using a partially-frozen encoder for VLAs could help retain pretrained representations. Instead of completely freezing visual encoders, which prior work~\cite{Kim2024openvla} has shown to degrade performance, we propose to use two siamese encoders: a \textit{frozen} one that serves as an ``anchor'' to retain the robust, semantically rich representations from VLM pretraining, and a \textit{finetuned} one that keeps the full flexibility to be specialized to robotic actions prediction. Mathematically,
\[
z_t = \big[\, \phi_{\text{frozen}}(o_t) \;\Vert\; \phi_{\text{train}}(o_t) \,\big],
\]  
where $\phi_{\text{frozen}}$ denotes the frozen encoder, $\phi_{\text{train}}$ the trainable encoder, and $\Vert$ vector concatenation. The combined latent representation $\mathbf{z}_t$ is then passed to a language-conditioned action tokenizer $\psi$ to generate an action token sequence,    
\[
a_t = \psi(z_t, c),
\] 

In practice, when the base VLA model provides only a single vision encoder, we duplicate it and freeze one copy (e.g.,~\cite{Black2024pizero}); if it already includes two, we freeze one while adapting the other (e.g.,~\cite{Kim2024openvla}).


\subsection{String-Based Tokenizer}



To maximally reuse pretrained language representations, we render each action dimension as a character sequence and predict it with the same autoregressive objective used in language pretraining. For example, the action component $\Delta x = 0.0312$ is tokenized into: 
\[
\begin{array}{cccccc}
\fbox{\makebox[1em][c]{\strut\texttt{0}}} &
\fbox{\makebox[1em][c]{\strut\texttt{.}}} &
\fbox{\makebox[1em][c]{\strut\texttt{0}}} &
\fbox{\makebox[1em][c]{\strut\texttt{3}}} &
\fbox{\makebox[1em][c]{\strut\texttt{1}}} &
\fbox{\makebox[1em][c]{\strut\texttt{2}}}
\end{array}
\]
Each box denotes a single character token from the language vocabulary. Representing numerical values as strings allows a single head to be co-trained on different prediction objectives (\ie, action for robot data and non-action prediction for vision-language datasets), and makes spatial information in pretrained language representations potentially more useful in adapting to robot action prediction, thereby improving generalization. Autoregressive string generation also refines actions step by step, yielding more precise predictions. We train our model with 4 decimals action.

Although string-based action tokenization along with the dual encoder together can increase model inference time by 0.5×–1.3× (depending on the model), we later show that the benefits substantially outweigh this cost.




\subsection{Feature Regularization via Co-training}

Learning solely from robot datasets often leads to overfitting, particularly in low-data regimes common in robotics. To counter this, we co-train VLAs using a mixture of robot data and vision-language datasets (\tabref{tab:datasets}), enabled by our shared string-based language and action representation. We utilize a mixture of spatial reasoning, spatial affordance, and general vision-language reasoning data. For every training batch, we sample 50\% from each type to balance the gradient. We hypothesize that such a joint training strategy prevents catastrophic forgetting and enhances generalization. 



\begin{table}[t]
\centering
\scriptsize
\caption{\textbf{Co-training datasets.}}
\renewcommand{\arraystretch}{1.1} 
\begin{tabular}{p{1.3cm} p{2.3cm} p{3.5cm}}
\toprule
\textbf{Type} & \textbf{Source} & \textbf{Description} \\
\midrule
Robot & OXE~\cite{OpenX} & Real robot demonstrations \\
\midrule
\multirow{7}{*}{\begin{tabular}{@{}l@{}}Vision-\\Language\\Reasoning\end{tabular}} & LLaVA Visual Instruct CC3M~\cite{sharma2018conceptual} & Image captioning and object grounding \\
& VQASynth-Spatial~\cite{chen2024spatialvlm} & Spatial reasoning via image--language queries \\
& LLaVA OneVision~\cite{li2024llava} & OCR, chart, and multimodal question answering \\
& RoboPoint~\cite{robopoint} & Spatial grounding of language into pixel locations \\
\bottomrule
\end{tabular}
\label{tab:datasets}
\vspace{-10pt}
\end{table}

\section{Experiments}

This section seeks to answer the following questions:
\begin{itemize}
    \item[\textbf{Q1.}] Are the proposed designs effective in learning robot manipulation policies? 
    \item[\textbf{Q2.}] Does our approach effectively preserve pretrained VLM representations, thereby improving generalization and robustness to novel visuals and instructions?  
    \item[\textbf{Q3.}] How well does the learned policy transfer to real-world environments? 
\end{itemize}

We investigate \textbf{Q1-Q2} in simulation and offline vision-language benchmark datasets for scalable and controlled study, and validate \textbf{Q3} on real-world robotic platforms.

For simulation, we use SimplerEnv~\cite{simplerenv}, which provides evaluation results that correlate well with real-world behaviors. To systematically study Q1–Q2, we evaluate under two SimplerEnv setups. The \textbf{Visual Matching} setup contains visuals closely matching the VLA training data, serving as an \textit{in-distribution} evaluation. The \textbf{Visual Variant Aggregation} setup introduces backgrounds, lightings, table textures, distractor objects, and camera poses out of the distribution of the robot data (OOD), serving as a generalization test.

\subsection{Analysis of Different Design Choices}

\noindent\textbf{Evaluation setup.}
We first perform SimplerEnv - Visual Matching evaluation to filter our design choices, as running the Visual Variant Aggregation evaluation is significantly more costly. We evaluate on four representative manipulation tasks, \pickcan{}, \opendrawer{}, \closedrawer{}, and \movenear{}, each over 300 episodes.

\noindent\textbf{Model setup.}
We utilize two representative VLA architectures: \openvla{} and \pizero{}, and apply our method from Sec.~\ref{sec:method}. For \pizero{}, which uses a flow-matching action head, we retain the original head for the baseline and replace it with string-based tokenizer for our method. All models, both baselines and ours, are initialized from their respective pretrained VLMs (DINO + SigLIP + Llama for \openvla{}; PaliGemma for \pizero{})
We then finetune the models on the RT-1 dataset \cite{brohan2022rt,OpenX}, optionally with co-training and our method.

\begin{table}[t]
\centering
\small
\caption{
\textbf{Ablation on SimplerEnv (Visual Matching).} The Visual Matching evaluation setting contains visuals highly similar to the robot action dataset used for VLA training. D, S, and C denote dual encoder, string tokenizer, and co-training, respectively. \ouropenvla{} and \ourpizero{} denote the corresponding models trained with all of D, S, and C.
}
\label{tab:ablation_designs}
\setlength{\tabcolsep}{4pt}
\begin{tabular}{
    lccccc
}
\toprule
\textbf{Model} &
\begin{tabular}{@{}c@{}}\texttt{Pick}\\\texttt{Can}\end{tabular} &
\begin{tabular}{@{}c@{}}\texttt{Open}\\\texttt{Drawer}\end{tabular} &
\begin{tabular}{@{}c@{}}\texttt{Close}\\\texttt{Drawer}\end{tabular} &
\begin{tabular}{@{}c@{}}\texttt{Move}\\\texttt{Near}\end{tabular} &
\textbf{Avg} \\
\midrule
OpenVLA         & 36.70 & 16.70 & 26.70 & 60.03 & 35.03 \\
OpenVLA+D       & 65.33 & 33.93 & 59.20 & 63.75 & 55.55 \\
OpenVLA+S       & 78.00 &  1.00 & 65.00 & 57.00 & 50.25 \\
OpenVLA+C       & 66.67 & \textbf{64.41} & 11.11 & 62.00 & 51.05 \\
OpenVLA+DS      & 90.00 & 44.00 & 72.10 & 70.00 & 69.03 \\
OpenVLA+SC      & 78.34 & 62.00 & 92.00 & \textbf{80.34} & 78.17 \\
OpenVLA$^{+}$   & \textbf{90.32} & 53.50 & \textbf{92.40} & 77.60 & \textbf{78.46} \\
\midrule
\pizero{} w/ Inst. Aug. \tablefootnote{Throughout the paper, we trained \pizero{} baselines with instruction augmentation, as omitting it led to substantially poorer performance; our \ouropenvla{} and \ourpizero{} models are trained without such augmentation. This setup is thus favorable to the baselines in comparison.}       & \textbf{84.33} & 30.40 & 57.29 & 78.54 & 62.64 \\
$\pi_{0}^{+}$   & 83.40 & \textbf{44.35} & \textbf{64.37} & \textbf{84.66} & \textbf{69.19} \\
\bottomrule
\end{tabular}
\end{table}


\begin{figure}[t]
    \centering
    \includegraphics[width=\linewidth]{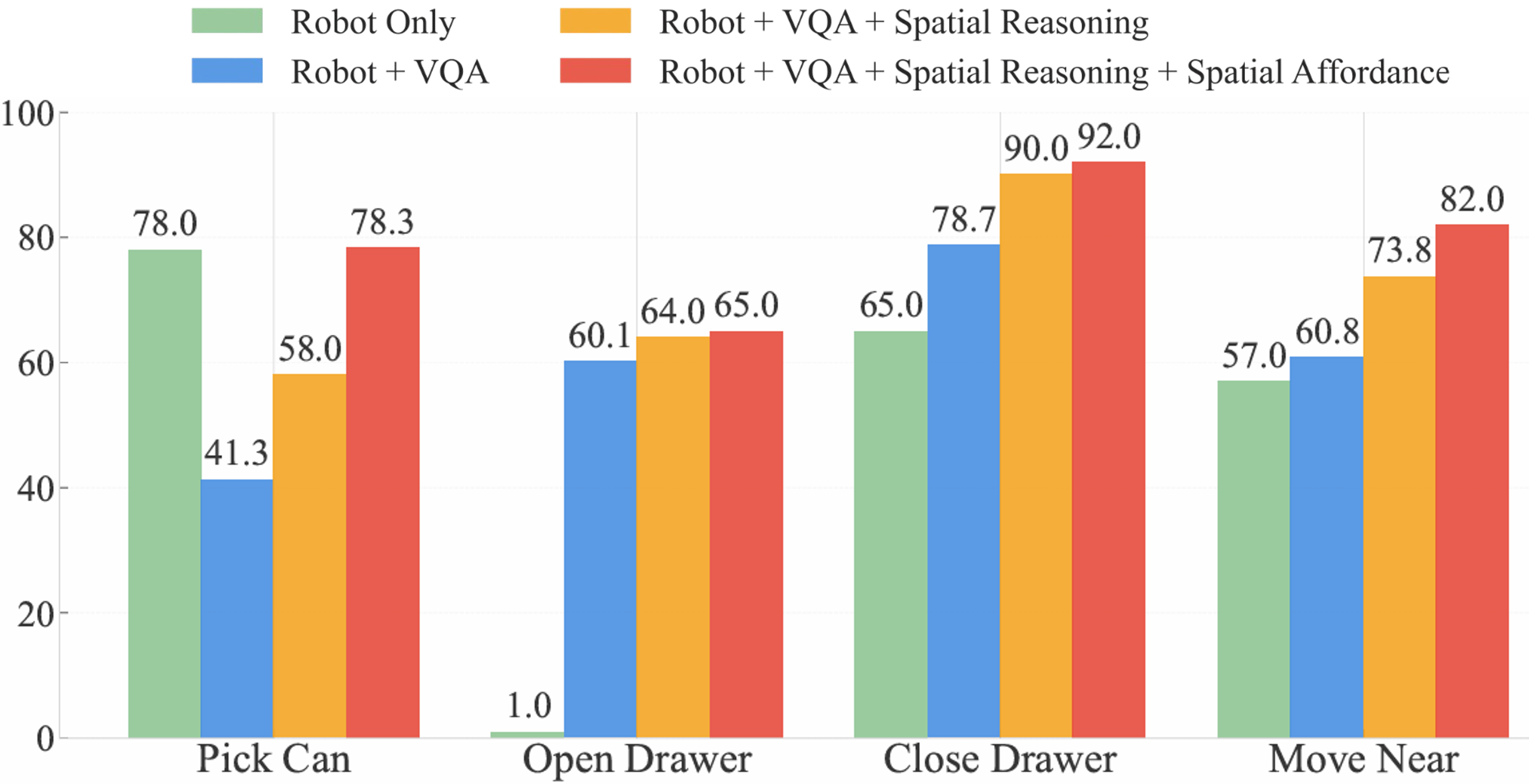}
    \caption{\textbf{Co-training data composition ablation on SimplerEnv (Visual Matching).} With our unified string-based tokenizer, combining diverse datasets for co-training consistently improves robotic task performance.}
    \label{fig:ablation_dataset}
\end{figure}

\noindent\textbf{Results.} We present the Visual Matching evaluation results in \tabref{tab:ablation_designs}. Overall, models trained with our recipe, \ouropenvla{} and \ourpizero{}, outperform \openvla{} and \pizero{} by nearly 40\% and 7\% respectively, showcasing the effectiveness of our method. 

\noindent\textbf{Effect of partially-frozen dual vision encoder.} Utilizing our partially-frozen dual encoder alone improves \openvla{} performance from 35.03\% to 55.55\%. When combined with our string tokenizer and co-training approaches, the performance further rises to 78.46\%, though the gain over only using String Tokenizer + co-training is marginal (78.17\% vs. 78.46\%). However, the benefits of dual encoder will become clear when we later assess the robustness and generalization of VLAs using vision and language perturbations.

\noindent\textbf{Effect of string tokenizer and vision-language co-training.}
Adding string tokenizer to \openvla{} alone raises the success rate from 35.03\% to 50.25\%. Moreover, when vision-language co-training is utilized, the model trained with string tokenizer performs significantly better than the one without (78.17\% vs. 51.05\%). 
This result indicates that unifying the action and language output space makes it easier to harness pretrained features, while not compromising the model's ability to adapt to precise robot actions. We also analyze the effect of data composition in co-training, showing that our selected datasets are well aligned with robotic tasks and consistently yield positive performance gains (\figref{fig:ablation_dataset}).



\subsection{Analysis of Representation Generalization and Robustness}

\begin{table*}[t]
\centering
\small
\caption{\textbf{Out-of-distribution (OOD) visual generalization evaluation in SimplerEnv}. ``Masked Background'' refers to Visual Matching evaluation but with background removed (illustrated in Fig.~\ref{fig:motivating}(a)). ``Visual Variant Aggregation'' follows the original evaluation protocol in SimplerEnv, randomizing backgrounds, table textures, lightings, distractors, and camera poses for testing OOD generalization. We also report OpenVLA+SC in this table to show that \ouropenvla{} has better robustness, despite performing very similarly for in-distribution evaluation in Tab.~\ref{tab:ablation_designs}. }
\label{tab:visual_variant}
\setlength{\tabcolsep}{6pt}
\begin{tabular}{
l
ccccc
ccccc
}
\toprule
\multirow{3}{*}{Model} & \multicolumn{5}{c}{Masked Background} & \multicolumn{5}{c}{Visual Variant Aggregation} \\
\cmidrule(lr){2-6}\cmidrule(lr){7-11}
 &
\begin{tabular}{@{}c@{}}\texttt{Pick}\\\texttt{Can}\end{tabular} &
\begin{tabular}{@{}c@{}}\texttt{Open}\\\texttt{Drawer}\end{tabular} &
\begin{tabular}{@{}c@{}}\texttt{Close}\\\texttt{Drawer}\end{tabular} &
\begin{tabular}{@{}c@{}}\texttt{Move}\\\texttt{Near}\end{tabular} &
Avg &
\begin{tabular}{@{}c@{}}\texttt{Pick}\\\texttt{Can}\end{tabular} &
\begin{tabular}{@{}c@{}}\texttt{Open}\\\texttt{Drawer}\end{tabular} &
\begin{tabular}{@{}c@{}}\texttt{Close}\\\texttt{Drawer}\end{tabular} &
\begin{tabular}{@{}c@{}}\texttt{Move}\\\texttt{Near}\end{tabular} &
Avg
\\
\midrule
OpenVLA   & 76.3 & 28.6 & 43.8 & 73.3 & 55.5  & 67.8 & 24.7 & 43.0 & 74.0 & 52.4 \\
OpenVLA+SC & 80.5 & 48.3 & 89.5 & 86.2 & 76.2 &82.3 & 42.3 & 81.3 & 86.6 & 73.1\\
OpenVLA$^{+}$ & 82.9 & 50.0 & 87.7 & 85.0 & \textbf{76.4}  & 87.6 & 48.4 & 87.9 & 82.3 & \textbf{76.6} \\
\midrule
\pizero{} w/ Inst. Aug.  & 86.3 & 25.0 & 42.8 & 75.2 & 57.3  & 77.0 & 32.3 & 38.6 & 65.0 & 53.2 \\
$\pi_0^{+}$   & 70.3 & 36.3 & 64.7 & 73.4 & \textbf{61.2}  & 60.3 & 35.5 & 60.5 & 83.1 & \textbf{59.9} \\
\bottomrule
\end{tabular}
\end{table*}

\begin{figure*}[t]
    \centering
    \includegraphics[width=\linewidth]{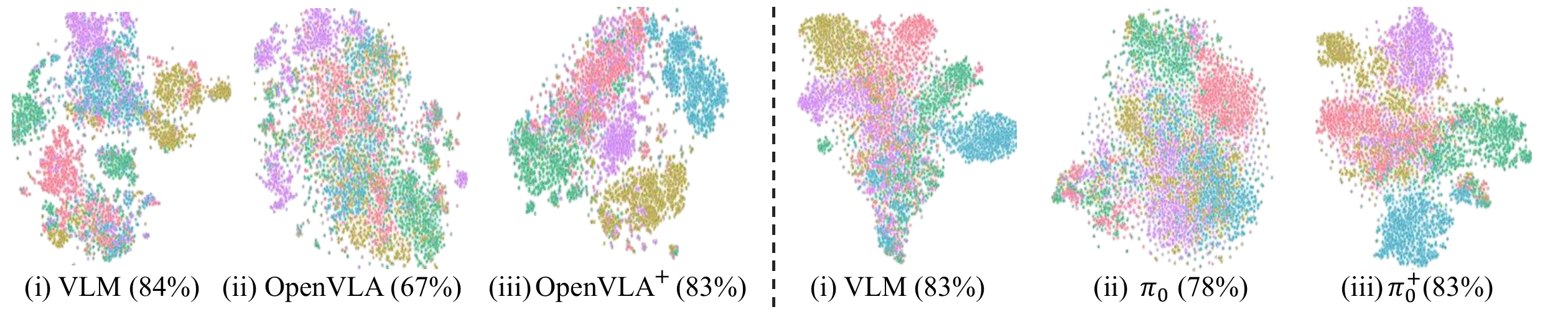}
    \vspace{-1.75em}
    \caption{ \textbf{t-SNE visualizations of vision encoder features on CIFAR-10.} We compare (i) original visual backbone from VLM before VLA training, (ii) after direct VLA fine-tuning on robot data, and (iii) after our approaches in Sec.~\ref{sec:method}. Numbers indicate linear-probe classification performance on CIFAR-10 using the corresponding features. For both \openvla{} and \pizero, our approach yields noticeably tighter, more well-separated class clusters that lead to better linear-probe performance, indicating better preservation of semantic structures in pretrained visual representations. }
    \label{fig:tsne_analysis}
    \vspace{-1em}
\end{figure*}

\begin{table}[t]
\centering
\caption{
\textbf{Language robustness evaluation.} Evaluating policies under original and paraphrased instructions. ``w/ Inst. Aug.'' indicates training with instruction augmentations. We also report OpenVLA+SC in this table to show that \ouropenvla{} has better robustness, despite performing very similarly for in-distribution evaluation in Tab.~\ref{tab:ablation_designs}.
}
\begin{tabular}{lcccc}
\toprule
\multirow{2}{*}{Model} & \multicolumn{2}{c}{\pickcan{}} & \multicolumn{2}{c}{\movenear{}} \\
\cmidrule(lr){2-3}\cmidrule(lr){4-5}
               & Orig. & Para. & Orig. & Para. \\
\midrule
OpenVLA & 36.70 & 12.12 & 60.03 & 42.07 \\
OpenVLA w/ Inst. Aug. & 30.56 & 38.89 & 56.49 & 50.00 \\
OpenVLA+SC & 78.34 & 62.52 & \textbf{80.34} & \textbf{82.10}\\
\ouropenvla{} & \textbf{90.32} & \textbf{84.52} & 77.60 & 78.19 \\

\midrule
\pizero{} w/ Inst. Aug. & \textbf{84.33} & 42.20 & 78.52 & 58.24 \\
\ourpizero{} & 83.40 & \textbf{55.40} & \textbf{84.66} & \textbf{63.33}\\

\bottomrule
\end{tabular}
\vspace{-15pt}
\label{tab:language_aug}
\end{table}

\noindent\textbf{Evaluation setup.}
We assess whether the learned visual and language representations in our models are rich, robust, and thus enable generalizable robot manipulation. We first evaluate \textbf{visual robustness} in two setups: (i) \textbf{background masking}, where background pixels are masked to test visual sensitivity, and (ii) \textbf{visual variant aggregation} in SimplerEnv \cite{simplerenv}, where task environments are visually randomized to assess generalization across environmental visual appearances. We also visualize learned representations on standard vision datasets for a qualitative analysis. For \textbf{language robustness}, we use GPT-4 to generate synonymous instructions (e.g., ``grasp the can'' and ``get the can'' for \pickcan{}), and evaluate zero-shot generalization to these unseen language variants. In addition, we evaluate our models on standard VQA benchmarks to analyze their representation quality through their general reasoning capabilities.

\noindent\textbf{Visual robustness.}
Both \ouropenvla{} and \ourpizero{} demonstrate substantial improvements in visual robustness across both background masking and visual variant aggregation. Table~\ref{tab:visual_variant} shows our methods outperform their respective baseline models on average across four robotic manipulation tasks. 

\noindent\textbf{Visualizing learned visual representations.} 
We extract learned representations on the CIFAR-10 dataset before and after VLA training, and visualize the low-dimensional embeddings obtained with t-SNE  (\figref{fig:tsne_analysis}). Embeddings of baseline VLAs typically exhibit unclear boundaries between different classes, while representations from our models are much more linearly separable, yielding much higher linear-probe classification performance. 


\noindent\textbf{Language robustness.} Table~\ref{tab:language_aug} compares three settings: (1) baseline VLA models, (2) baselines trained with paraphrased-instruction augmentation, and (3) our models. 
Unlike (2), where the paraphrased instructions used for evaluation are disjoint from those seen during training, both (1) and (3) have not seen any paraphrased instructions. The same set of paraphrased instructions is used consistently across all evaluations.
Despite receiving no language augmentations, our models substantially outperform baselines trained with augmentation. This shows that our method achieves stronger instruction generalization by leveraging pretrained representations through our string-based action tokenizer, effectively inheriting the language understanding and spatial reasoning capabilities of upstream VLMs.


\begin{figure}[t]
    \centering
    \includegraphics[width = 0.8\columnwidth]{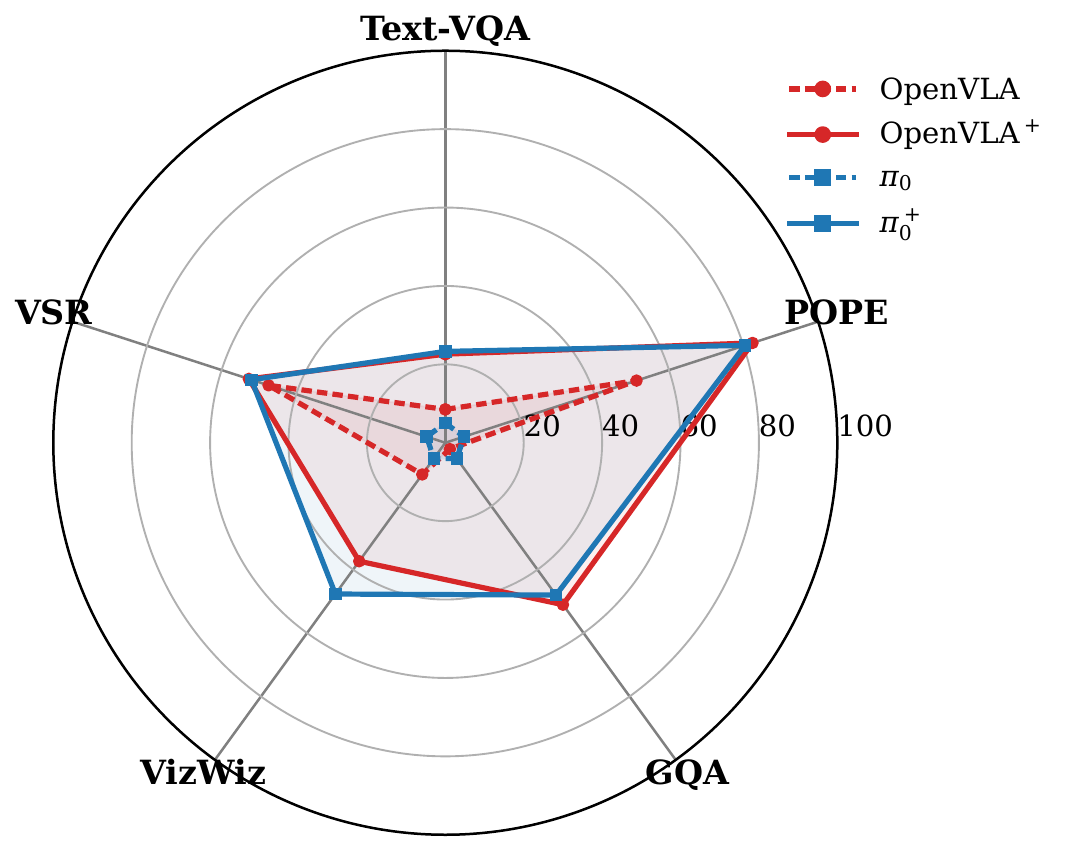} 
    \caption{\textbf{Evaluating VLAs on five VQA benchmarks.} \ouropenvla{} and \ourpizero{} achieve significantly higher performance across all tasks, showing that our training recipe helps retain pretrained representations of VLM backbones.}
    \label{fig:vla_on_vqa_eval}
    \vspace{-15pt}
\end{figure}

\noindent\textbf{Reasoning abilities.}
To further analyze the representation quality of our models, we present results on common VLM benchmarks like Text-VQA  \cite{singh2019towards}, POPE \cite{li2023evaluating}, GQA \cite{hudson2019gqa}, VizWiz\cite{bigham2010vizwiz}, and VSR\cite{liu2023visual}  and report the accuracy metric in Fig.~\ref{fig:vla_on_vqa_eval}. These benchmarks include many challenging reasoning questions that place high demands on the underlying representations. We find that baseline \openvla{} and \pizero{} perform poorly, whereas \ouropenvla{} and \ourpizero{} achieve substantially higher accuracy, demonstrating that our training recipe more effectively preserves the reasoning capabilities of pretrained VLM representations within VLAs.

\begin{table*}[t]
\centering
\begin{tabular}{lccccc}
\toprule
\textbf{Task} & \textbf{Distractor} & \textbf{OpenVLA} & \textbf{\ouropenvla{}} & \textbf{\pizero{} w/ Inst. Aug.} & \textbf{\ourpizero{}} \\
\midrule
\pickcarrot{}            & None & 32.0 & \textbf{60.0} & 16.0 & 40.0 \\
\pickeggplant{}         & None & 8.0  & \textbf{24.0} & 8.0  & \textbf{24.0} \\
\pickknife{}             & None & 36.0 & \textbf{64.0} & 24.0 & 40.0 \\
\placeknife{}  & None & 0.0  & 12.0 & 8.0  & \textbf{20.0} \\
\putknifecloth{}     & None & 4.0  & 12.0 & 12.0 & \textbf{48.0} \\
\putcarrotplate{}    & None & 4.0  & 16.0 & 12.0 & \textbf{36.0} \\
\placecornknife{}  & None & 0.0  & 0.0  & 0.0  & 0.0 \\
\midrule
\pickcarrot{} & Knife & 20.0 & \textbf{48.0} & 12.0 & 36.0\\
\pickknife{} & Carrot & 12.0 & \textbf{36.0} & 8.0 & 28.0\\
\putcarrotplate{} & Knife & 8.0 & 12.0 & 8.0 & \textbf{36.0} \\
\putcarrotcolorplate{} & Plate  & 0.0 & 8.0 & 0.0 & \textbf{20.0}\\
\putcarrotplate{} & Cloth & 4.0 & 8.0 & 4.0 & \textbf{32.0} \\
\bottomrule
\end{tabular}
\caption{\textbf{Real-world evaluation results.} Distractors are irrelevant objects that may mislead the policy, such as similar items (e.g., differently colored plates) or dissimilar ones (e.g., knife, carrot, cloth), and are used to assess the robustness of the policies. Our models consistently outperform baselines in their presence.}
\label{tab:real-eval-results}
\end{table*}

\subsection{Real-World Evaluation}

\noindent\textbf{Evaluation setup.}
We initialized VLAs from their respective pretrained VLMs and finetune the policies on the Bridge dataset. We deploy both baselines and our method on the ViperX 300s
\footnote{Although a WidowX setup was not available, we adapted the ViperX control stack to closely match that of the WidowX.}
robot to evaluate real-world performance. Our experimental configuration utilizes a Logitech C920 Webcam for visual input \cite{Walke2023BridgeDataV2} and dual GTX 1080 Ti GPUs
\footnote{More advanced GPUs were not available in the room where we conducted the experiments. However, this should not affect our findings.}
. We evaluate models on both in-distribution and out-of-distribution (OOD) instructions, along with potentially unseen distractors, to comprehensively analyze model capabilities. We conduct 25 trials per task and per model. For a 7-DoF action, the inference time is about 1.53s for baseline \openvla{}, 2.30s for \ouropenvla{}, 0.73s for baseline \pizero{}, and 1.70s for \ourpizero{}.

\begin{figure}[t]
    \centering
    \includegraphics[width=\linewidth]{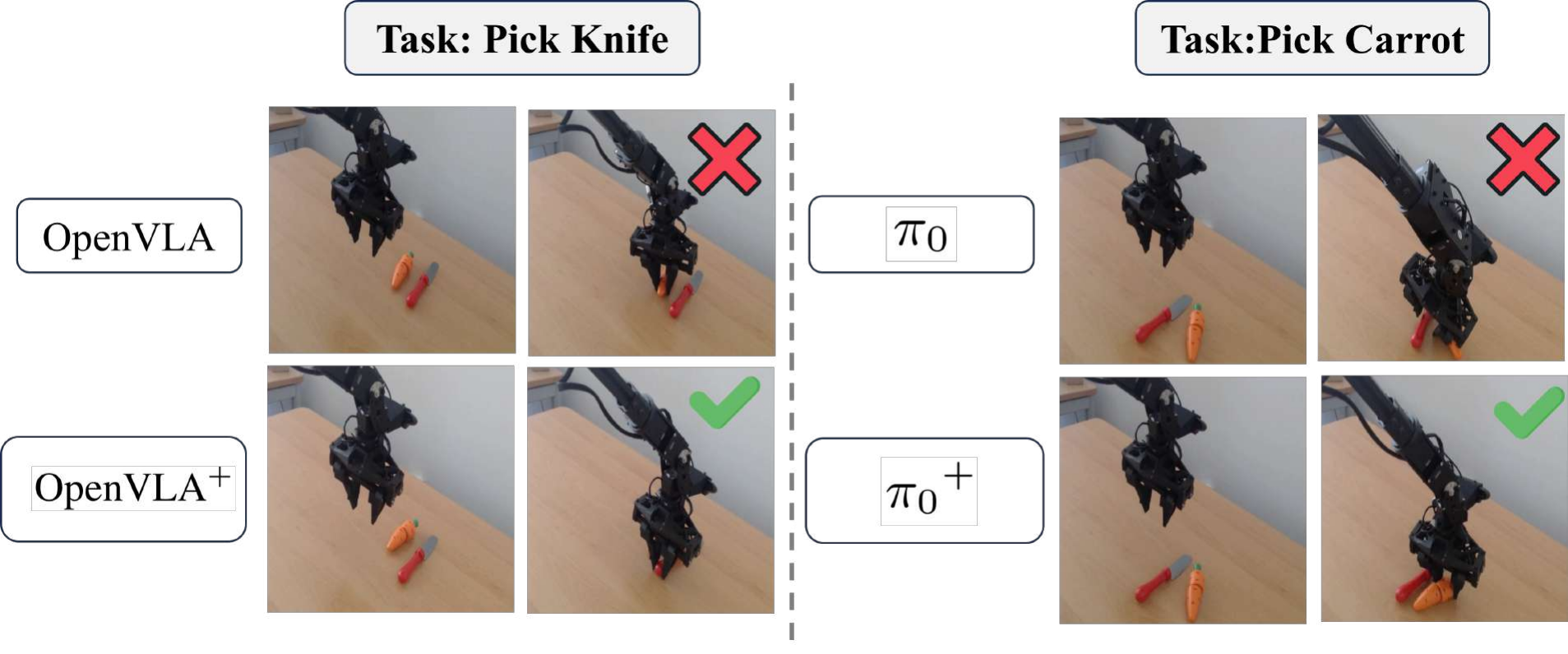}
    \caption{\textbf{Qualitative results on the \pickknife{} and \pickcarrot{} tasks.} Our models have stronger robustness to distractors.}
    \label{fig:Real_eval}
    \vspace{-10pt}
\end{figure}

\noindent\textbf{Results.} \figref{fig:Real_eval} and Table~\ref{tab:real-eval-results} illustrate the results. We find that our models consistently outperform baseline VLAs on all manipulation tasks. In particular, we observe that base models such as OpenVLA and \pizero{} tend to misinterpret the task under distractors, often reaching for the closest object in the scene rather than the instructed target, e.g., picking a carrot instead of a knife in \pickknife{}, or failing to pick the correct carrot among other similar items. In contrast, our approach demonstrates a more robust understanding of the task, successfully completing goal-conditioned actions even in the presence of semantically or spatially similar distractors, and achieving substantially better performance.

\section{Conclusion}
In this paper, we proposed a set of approaches that improve the robustness and generalization of vision-language-action (VLA) models by better preserving the robust representation structures from pretrained vision-language models (VLMs). These include a dual-visual-encoder design that mixes pretrained and finetuned features, aligning robotic action with language output via a string-based tokenizer to better transfer pretrained knowledge, and a balanced co-training approach on both robot and vision-language reasoning data. Experiments in both simulation and the real world demonstrate that our designs enable better learning of robot manipulation policies, better generalization and robustness to novel visuals and instructions, and better performance in the real-world.




\printbibliography





\end{document}